\documentclass[10pt,twocolumn,letterpaper]{article}

\usepackage{cvpr}             
\usepackage{times}
\usepackage{caption}
\usepackage{epsfig}
\usepackage{graphicx}
\usepackage{amsmath}
\usepackage{amssymb}
\usepackage{bm}
\usepackage{xcolor}
\usepackage{array}
\usepackage{booktabs}
\usepackage{multirow}
\usepackage[accsupp]{axessibility}
\usepackage{lineno}
\usepackage{makecell}
\usepackage{pifont}
\newcommand{\cxmark}{\ding{55}}
\newcommand{\ccheckmark}{\ding{51}}

\definecolor{mygreen}{HTML}{008000}
\definecolor{myred}{HTML}{D10000}

\definecolor{cvprblue}{rgb}{0.21,0.49,0.74}
\usepackage[pagebackref,breaklinks,colorlinks,citecolor=cvprblue]{hyperref}

\def\paperID{10373} 
\def\confName{CVPR}
\def\confYear{2025}

\title{FFHQ-Makeup: Paired Synthetic Makeup Dataset \\with Facial Consistency Across Multiple Styles}

\author{Xingchao Yang$^{1}$, Shiori Ueda$^{2}$, Yuantian Huang$^{1}$, Tomoya Akiyama$^{1}$, Takafumi Taketomi$^{1}$\\
$^1$CyberAgent \quad$^2$Keio University\\
}

\begin{document}

\maketitle

\begin{abstract}
Paired bare-makeup facial images are essential for a wide range of beauty-related tasks, such as virtual try-on, facial privacy protection, and facial aesthetics analysis. However, collecting high-quality paired makeup datasets remains a significant challenge. Real-world data acquisition is constrained by the difficulty of collecting large-scale paired images, while existing synthetic approaches often suffer from limited realism or inconsistencies between bare and makeup images.
Current synthetic methods typically fall into two categories: warping-based transformations, which often distort facial geometry and compromise the precision of makeup; and text-to-image generation, which tends to alter facial identity and expression, undermining consistency.
In this work, we present \textit{\textbf{FFHQ-Makeup}}, a high-quality synthetic makeup dataset that pairs each identity with multiple makeup styles while preserving facial consistency in both identity and expression. Built upon the diverse FFHQ dataset, our pipeline transfers real-world makeup styles from existing datasets onto 18K identities by introducing an improved makeup transfer method that disentangles identity and makeup. Each identity is paired with 5 different makeup styles, resulting in a total of 90K high-quality bare–makeup image pairs.
To the best of our knowledge, this is the first work that focuses specifically on constructing makeup dataset.
We hope that FFHQ-Makeup fills the gap of lacking high-quality bare–makeup paired datasets and serves as a valuable resource for future research in beauty-related tasks.
\end{abstract}


\section{Introduction}
Makeup plays a multifaceted role in human appearance, influencing not only facial aesthetics but also perceptions of identity, personality, and even social behavior. In the context of computer vision, the ability to analyze, manipulate, and transfer makeup styles has drawn increasing attention, enabling applications such as virtual try-on (VTO)~\cite{stablemakeup2024, makeupRecommend2017, ssat, beautyGAN}, face recognition under makeup variations~\cite{YMU_VMU_2012, MIW_2013, makeupfacerecog2018}, facial privacy protection~\cite{protectMakeup2022, antimakeup2018}, and beauty assessment~\cite{beauty3Dfacenet2021}. While the field has made remarkable progress in recent years, a critical bottleneck remains: the lack of open-source, large-scale, high-quality paired makeup datasets containing both bare and makeup images. This limitation hinders the development of robust and generalizable models, and significantly impedes progress in makeup-related applications.

A well-constructed paired makeup dataset is expected to meet several key requirements. First, it must maintain high makeup realism, ensuring that the applied styles are both plausible and visually convincing. Second, it should exhibit diversity in both facial identities and makeup styles to reflect real-world variability. Third, it should ensure facial consistency across pairs, preserving identity and facial structure despite the presence of makeup. In addition, the dataset should ideally be large enough to support the data requirements of modern deep learning models.

\begin{figure}[t]
    \centering
    \includegraphics[width=\linewidth]{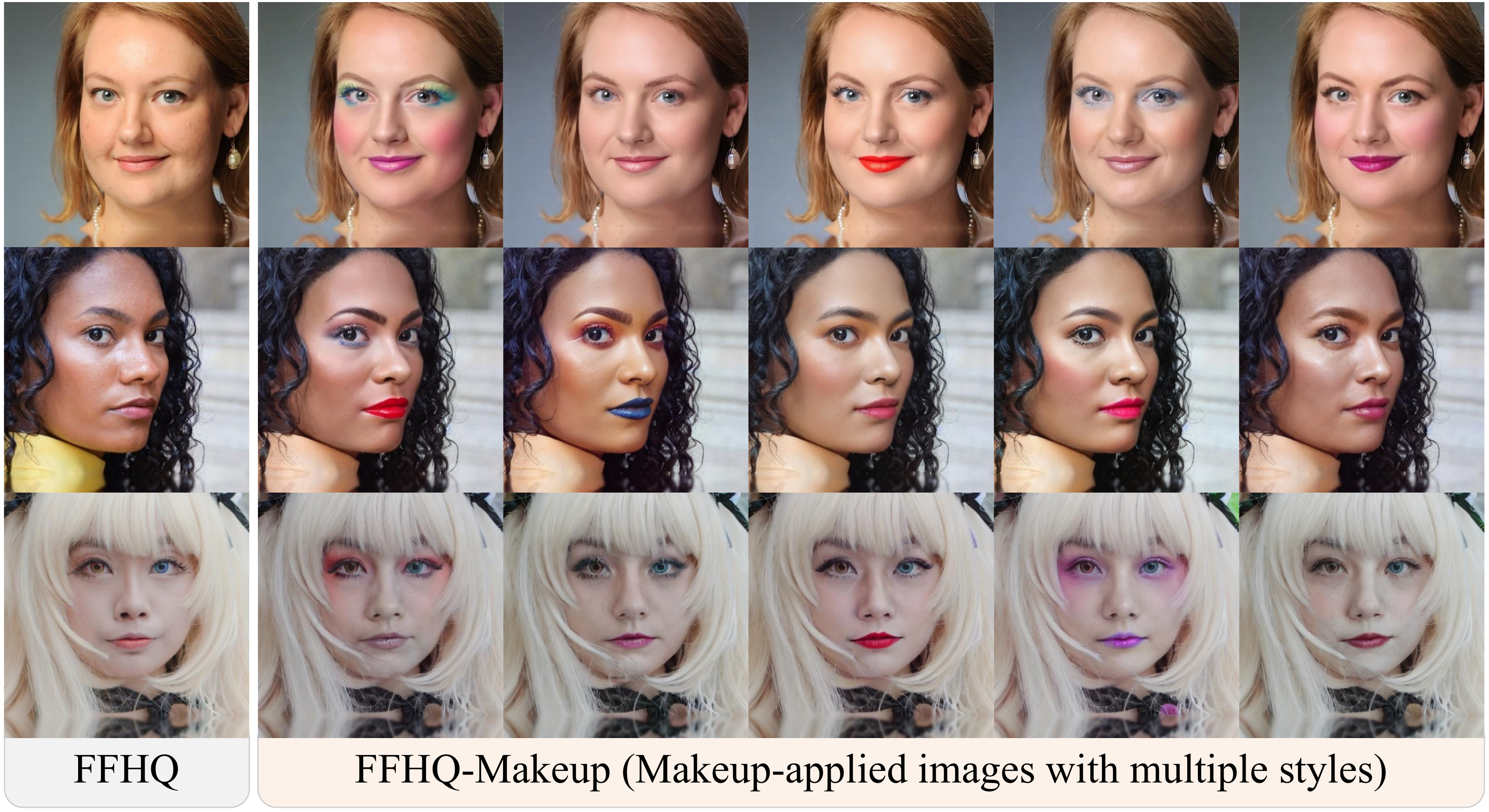}
    \caption{Examples from our FFHQ-Makeup dataset, where each identity is paired with a bare image and multiple makeup-applied images, demonstrating identity and expression consistency under diverse makeup styles.}
\label{fig:teaser}
\end{figure}

\begin{table*}
    \caption{Summary of existing makeup datasets.}
    \begin{center}
    \resizebox{\linewidth}{!}{
    {\small
    \begin{tabular}{@{}lrrrrrrrrr@{}}
        \toprule
        \textbf{\makecell{Datasets}} & 
        \textbf{\makecell{Subjects}} &
        \textbf{\makecell{Images per\\Subject}} &
        \textbf{\makecell{Makeup\\Images}} & \textbf{\makecell{Non-makeup\\Images}} & \textbf{\makecell{Resolution}} & 
        \textbf{\makecell{Type}} & \textbf{\makecell{Paired}} & \textbf{\makecell{Public\\avail.}} \\
        \midrule
        YMU~\cite{YMU_VMU_2012} & 151 & 4 & 302 & 302 & 130 $\times$ 150 & Real & \ccheckmark & \cxmark\\
        MIW~\cite{MIW_2013} & 125 & 1-2 & 77 & 77 & -- & Real & \cxmark & \cxmark\\
        MIFS$^1$~\cite{MIFS_Chen_2017} & 214 & 2 or 4 & 214 & 428 & -- & Real & \ccheckmark & \cxmark\\
        FAM~\cite{FAM_Hu_2013} & 519 & 2 & 519 & 519 & 64 $\times$ 64 & Real & \ccheckmark & \cxmark\\
        MT~\cite{beautyGAN} & 2719 & 1-2 & 2719 & 1115 & 361 $\times$ 361 & Real & \cxmark & \ccheckmark\\
        LADN~\cite{ladn} & 635 & 1 & 302 & 333 & $\approx 320 \times 320$ & Real & \cxmark & \ccheckmark\\
        Wild~\cite{PSGAN} & 772 & 1 & 403 & 369 & 256 $\times$ 256 & Real & \cxmark & \ccheckmark \\
        CPM-Real~\cite{CPM} & 2895 & 1 & 2895 & -- & -- & Real & \cxmark & \ccheckmark \\
        BeautyFace~\cite{BeautyREC} & 44 & 1+ & 3000 & -- & 512 $\times$ 512 & Real & \cxmark & \ccheckmark \\
        \midrule
        VMU~\cite{YMU_VMU_2012} & 51 & 4 & 153 & 51 & 130 $\times$ 150 & Synthetic (Manually edited) & \ccheckmark & \cxmark\\
        LADN-Syn~\cite{ladn} & 333 & 355 & 120K & 333 & $\approx 320 \times 320$ & Synthetic (Warp-Paste) & \ccheckmark & \ccheckmark \\        
        Stable-Makeup~\cite{stablemakeup2024} & 20K & 1 & 20K & 20K & 512 $\times$ 512 & Synthetic (Text-to-Image$^2$) & \ccheckmark & \cxmark \\
        BeautyBank~\cite{BeautyBank_Lu_2025} & 70K & 1+ & 324K & 70K & 512 $\times$ 512 & Synthetic (Text-to-Image$^2$) & \ccheckmark & \ccheckmark \\
        \midrule
        \textit{\textbf{FFHQ-Makeup (Ours)}} & 18K & 5 & 90K & 18K & 512 $\times$ 512 & Synthetic (Generation) & \ccheckmark & \ccheckmark\\
        \bottomrule
    \end{tabular}}}
    \end{center}
    \scriptsize $^1$ For MIFS, makeup images = 214 (imposters only), non-makeup images = 214 (imposters) + 214 (targets) = 428 in total.
    
    \scriptsize$^2$ Stable-Makeup uses LEDITS~\cite{ledits2023} for makeup synthesis, while BeautyBank adopts the same strategy with improved LEDITS++~\cite{leditsplus2024}.
    \label{tab:datasets_analysis}
\end{table*}

\begin{figure}[t]
    \centering
    \includegraphics[width=\linewidth]{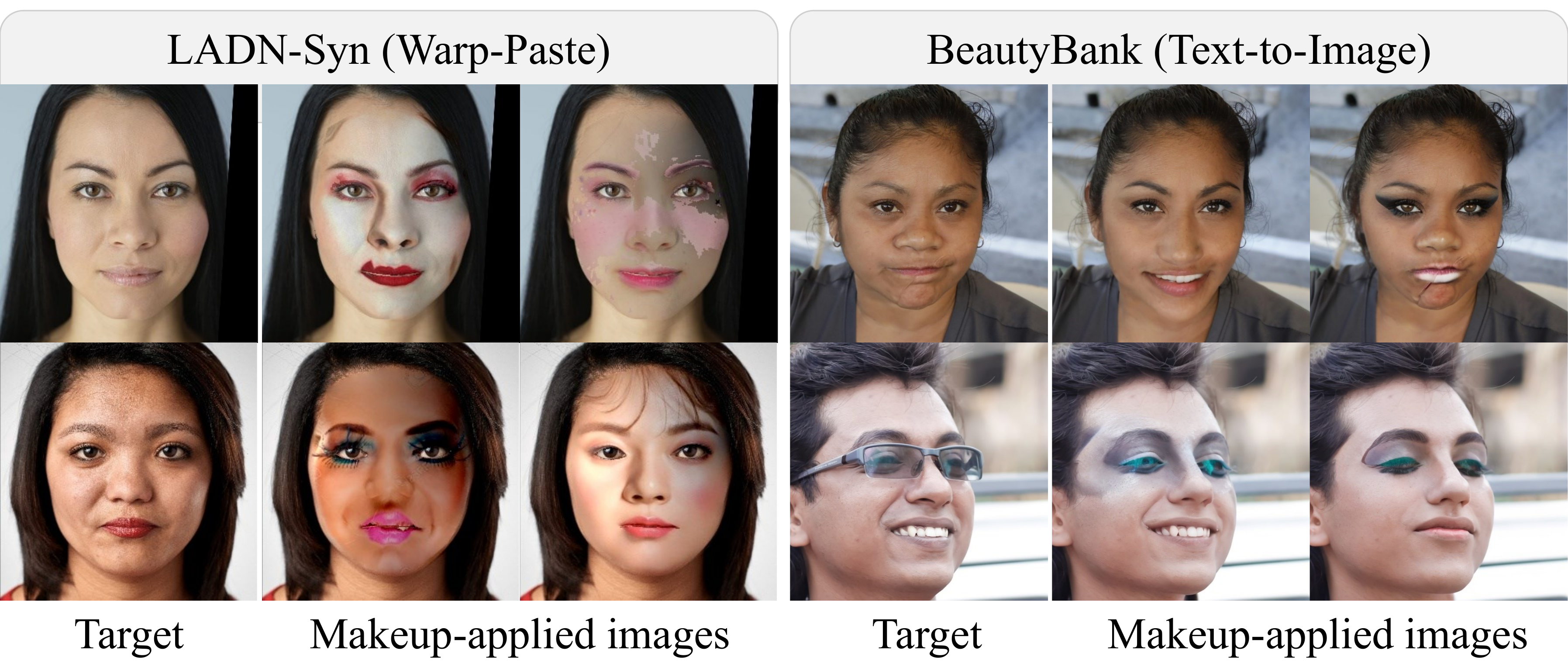}
    \caption{Examples of existing large-scale synthetic paired bare–makeup datasets. Existing methods often introduce artifacts or alter the identity and expression of the subject.}
\label{fig:dataset_compare}
\end{figure}

Despite ongoing efforts, existing makeup datasets construction still fail short of key requirements, as summarized in Tab.~\ref{tab:datasets_analysis}. Real-world data collection is hindered by logistical constraints, limited resources, and privacy concerns. As a result, these datasets are typically either limited in scale~\cite{YMU_VMU_2012, MIFS_Chen_2017, FAM_Hu_2013} or lack paired bare and makeup images~\cite{beautyGAN, ladn, PSGAN, CPM, BeautyREC}. 
To address these issues, synthetic approaches have recently gained popularity. However, as illustrated in Fig.~\ref{fig:dataset_compare}, they still suffer from various limitations: warping-based methods~\cite{ladn} often produce low-quality results and introduce facial distortions, while text-to-image generation~\cite{stablemakeup2024, BeautyBank_Lu_2025} frequently causes identity drift and inconsistencies in facial expression, and struggles to capture fine-grained makeup details due to the inherent ambiguity of language in describing subtle color gradients, eyeshadow geometry, and cheek contours. Consequently, existing resources offer limited utility for makeup-related applications.

To address these challenges, we present \textit{\textbf{FFHQ-Makeup}}: an open-source, large-scale, high-quality multiple paired synthetic makeup dataset designed to overcome the limitations of prior work.
We introduce a novel data generation method built upon the state-of-the-art makeup transfer method~\cite{stablemakeup2024}, with key improvements in both facial structure control and makeup feature extraction.
For structure control, our method eliminates the reliance on paired bare-makeup images, which are difficult to obtain in real-world scenarios. Instead, given a single makeup image, we employ a 3D Morphable Model (3DMM) fitting to reconstruct an approximate bare face counterpart. 
The fitted 3DMM not only captures the subject's identity, expression, pose, and skin tone, but also the illumination. 
This self-supervised approach enables scalable and flexible for synthetic makeup dataset generation.
For makeup feature extraction, we introduce a 3DMM-based residual representation of makeup appearance, combined with sampling and re-rendering augmentation strategies.
These techniques extend limited makeup data across diverse facial variations in FFHQ~\cite{FFHQ}, helping disentangle facial structure from makeup appearance and facilitating the generation of semantically consistent bare–makeup pairs.
During dataset construction, we manually filtered out failed cases and samples with visual artifacts to ensure data quality. The final FFHQ-Makeup dataset comprises 18K identities derived from FFHQ, each paired with 5 distinct makeup styles, resulting in a total of 90K paired images, as illustrated in Fig.~\ref{fig:teaser}.
We evaluate the effectiveness of both our dataset and method through extensive experiments and comparisons. 
We anticipate that this high-quality, multi-style paired dataset will greatly benefit a wide range of future makeup-related research and applications.

Our main contributions are as follows:
\begin{itemize}
\item \textbf{FFHQ-Makeup Dataset:}
We introduce a large-scale synthetic dataset with 18K identities and 5 diverse makeup styles per identity, providing 90K high-quality paired bare–makeup images.

\item \textbf{Pair-free Structure Control:}
We propose a scalable generation pipeline that reconstructs bare faces from single makeup images via 3DMM fitting, removing the need for paired supervision.

\item \textbf{Decoupled Makeup Synthesis:}
We extract residual makeup features based on 3DMM, then transfer them across diverse faces using sampling and re-rendering augmentation, effectively disentangling facial structure from makeup appearance.
\end{itemize}

\section{Related works}
\subsection{Makeup Datasets and Tasks}
We summarize representative makeup-related datasets in Tab.~\ref{tab:datasets_analysis}. Early studies primarily focused on face recognition and verification tasks under makeup conditions~\cite{YMU_VMU_2012, MIW_2013, FAM_Hu_2013, MIFS_Chen_2017, makeupfacerecog2018}, where small-scale, low-resolution datasets were collected to support experimental evaluations.
In recent years, the scope of makeup-related research has significantly expanded, including tasks such as facial privacy protection~\cite{protectMakeup2022, antimakeup2018, FM2u-Net_2020}, beauty assessment~\cite{beauty3Dfacenet2021}, makeup recommendation~\cite{facebeautify_2022, MakeupRecommend_2017}, and 3D facial makeup~\cite{bareskinnet, MakeupExtract, makeupPriors2024, GeneAvatar_2024}.
Makeup transfer~\cite{beautyGAN, ladn, BeautyGlow, PSGAN, SOGAN, SCGAN, CPM, ssat, elegant, BeautyREC, CSD-MT2024, BeautyBank_Lu_2025, TinyBeauty2024, SHMT2024, stablemakeup2024}, in particular, has emerged as the dominant research focus, with deep learning approaches increasingly requiring large-scale, high-quality training data.

Makeup transfer aims to apply the makeup style of a reference image to a target face, while keeping the target’s identity, pose, and expression unchanged. Achieving this goal is particularly difficult due to the lack of ground-truth training pairs.
Early works adopted GAN-based pipelines. BeautyGAN~\cite{beautyGAN} utilized a dual input/output architecture with a color histogram loss to guide region-wise color matching. To train the model, they collected a non-paired dataset (MT) with 2,719 makeup and 1,115 non-makeup images.
LADN~\cite{ladn} introduced multiple local discriminators to better handle heavy makeup styles, collecting 302 makeup and 333 non-makeup images. To augment data, they generated 120K synthetic samples (LADN-Syn) using warp-and-paste, though the realism remains limited.
PSGAN~\cite{PSGAN} addressed spatial misalignment via an attention-based makeup projection module, using a dataset of 403 makeup images (mostly side profiles) for evaluation.
CPM~\cite{CPM} tackled both color and pattern transfer via UV map representations and synthetic pattern datasets. They also collected a 2,895-image real makeup dataset (CPM-Real).
Compared to previous datasets, BeautyREC~\cite{BeautyREC} increased resolution to $512 \times 512$ and included 3,000 makeup images (BeautyFace).

Recently, diffusion-based approaches have enabled more flexible and fine-grained makeup transfer.
Stable-Stable-Makeup~\cite{stablemakeup2024} proposed a detail-preserving encoder and incorporated cross-attention layers. They also built a pseudo-parallel dataset of 20K image pairs for training, generated via GPT-4-guided editing~\cite{ledits2023} and refined through manual quality filtering. Unfortunately, the dataset is not publicly available.
Inspired by Stable-Makeup, BeautyBank~\cite{BeautyBank_Lu_2025} adopted an improved LEDITS++~\cite{leditsplus2024} strategy to generate a larger dataset. However, due to the lack of post-filtering, the resulting makeup and non-makeup images are often misaligned.

To compensate for the scarcity of high-quality dataset, we propose FFHQ-Makeup to fill a critical gap by offering a publicly available, high-quality, multi-reference paired makeup dataset that supports more robust and scalable research in makeup-related tasks.

\subsection{Facial Attributes with 3DMM}
Since its introduction by Blanz and Vetter~\cite{Blanz3DMM99}, the 3D Morphable Model (3DMM) has been widely used in a variety of face-related tasks~\cite{egger20203d}. It represents facial shape and texture as linear combinations of basis components learned from a collection of 3D faces.
Through 3DMM fitting~\cite{MonocularRTA}, it is possible to recover 3D facial attributes—such as identity, expression, and pose—from a single 2D image. These model-based attributes have proven effective for controllable image synthesis and facial attribute editing, allowing structured manipulation of identity, expression, texture, pose, and illumination~\cite{DiffSwap2023, ControlFace_2025, DiffusionRig, DiFaReli}.

In this work, we leverage the FLAME model~\cite{FLAME} to reconstruct a bare face representation from a given makeup image. This enables a synthetic pipeline that no longer relies on paired data, facilitating more flexible dataset generation.

\subsection{Diffusion Models for Face Editing}
Diffusion models~\cite{ddpm} are generative models that iteratively transform random noise into realistic images through a sequence of denoising steps.
Pretrained latent diffusion models, such as Stable Diffusion~\cite{stablediffusion2022}, have demonstrated strong capabilities in photorealistic image synthesis and controllable image editing. Building on this foundation, recent works have extended Stable Diffusion for face-related applications~\cite{ipadapter2023, instantid2024, PhotoMaker_2024}.

Stable-Makeup~\cite{stablemakeup2024} is the first work to introduce pre-trained Stable Diffusion into the makeup transfer task, setting a new standard in the field. Its architecture consists of two main components:
(1) a feature extraction module that employs the CLIP image encoder~\cite{CLIP2021}, aggregating multi-layer features from the visual backbone to capture fine-grained details. Additionally, it incorporates a self-attention-based mapping to more efficiently extract and align makeup features;
(2) a structure control module based on ControlNet~\cite{controlnet23}, which enables conditional generation guided by specific structural inputs without altering the basic diffusion model.
To further improve facial feature extraction, FreeUV~\cite{freeuv_yang_2025} builds on Stable-Makeup by introducing a channel-attention mapping mechanism, which enhances feature discrimination while mitigating the spatial interference introduced by self-attention.

Our method builds upon Stable-Makeup, incorporating the improved feature extraction design from FreeUV to enable robust and reliable makeup dataset generation.

\section{Approach: FFHQ-Makeup}

We propose that a high-quality paired makeup dataset should satisfy the following three properties:

\begin{itemize}
    \item \textbf{Makeup Realism} ($\mathcal{P}_\text{makeup}$): The makeup should appear natural and realistic in terms of texture, color, and spatial placement (e.g., lipstick, eyeshadow, blush), closely resembling real-world cosmetic applications.

    \item \textbf{Facial Diversity} ($\mathcal{P}_\text{diversity}$): The dataset should cover a wide range of facial identities and attributes.

    \item \textbf{Facial Consistency} ($\mathcal{P}_\text{consistency}$): Each bare-makeup image pair should preserve consistent underlying facial attributes except for the makeup itself.
\end{itemize}

However, satisfying all three properties simultaneously remains a significant challenge.  
Real-world makeup datasets~\cite{beautyGAN, ladn, CPM, YMU_VMU_2012, PSGAN} typically offer high makeup realism ($\mathcal{P}_\text{makeup}$), but collecting paired images with consistent pose, lighting, and expression is costly and scale-limited. Consequently, they often lack subject diversity and pairwise consistency ($\neg \mathcal{P}_\text{diversity}$, $\neg \mathcal{P}_\text{consistency}$).  
In contrast, synthetic datasets~\cite{ladn, stablemakeup2024, BeautyBank_Lu_2025} enable large-scale pair generation and broader identity coverage, but often fail to produce realistic makeup ($\neg \mathcal{P}_\text{makeup}$) or to maintain facial consistency across pairs ($\neg \mathcal{P}_\text{consistency}$).

To address these limitations, we develop an improved makeup transfer approach and apply it to transplant real-world makeup styles onto a diverse set of subjects from FFHQ~\cite{FFHQ}.
This strategy leverages authentic cosmetic styles to preserve makeup realism, while substantially enhancing subject diversity to better reflect real-world variability.
Due to limited subject diversity ($\mathcal{P}_\text{diversity}$) in real-world makeup sources, our method may occasionally fail to perfectly replicate every fine-grained detail of the original makeup.
However, rather than pursuing for pixel-perfect makeup reproduction, our goal and key innovation lie in creating a dataset that better fulfills the key criteria of $\mathcal{P}_\text{makeup}$, $\mathcal{P}_\text{diversity}$, and $\mathcal{P}_\text{consistency}$.

This section is organized in three parts: (1) data preparation, (2) method description, and (3) final dataset construction.

\subsection{Data Preparation}
\label{sec:data_preparation}
We use existing makeup datasets (MT~\cite{beautyGAN} and LADN~\cite{ladn}) as the makeup style source $\mathcal{S}$ to ensure high makeup realism ($\mathcal{P}_\text{makeup}$), and employ FFHQ as the target identity set $\mathcal{T}$ to ensure facial diversity ($\mathcal{P}_\text{diversity}$).
The resulting transferred dataset, denoted as $\mathcal{G}$, is constructed by applying makeup styles from $\mathcal{S}$ onto subjects from $\mathcal{T}$, and is expected to preserve facial consistency ($\mathcal{P}_\text{consistency}$) with respect to $\mathcal{T}$.

During training, we exclusively use the makeup dataset $\mathcal{S}$. For each makeup image $I^\mathcal{S} \in \mathcal{S}$, we obtain the following data: a facial mask $\mathcal{M}$, detected 2D landmarks $\mathcal{L}$, and a reconstructed 3D face $\mathcal{F}$ via 3DMM fitting~\cite{MakeupExtract}. Using the facial mask $\mathcal{M}$, we blend the reconstructed face $\mathcal{F}$ with the background of $I^\mathcal{S}$ to produce a reconstructed bare face $\hat{I}_\text{b}$. By subtracting this bare face from the original image, we derive the makeup residual $\mathcal{R} = I^\mathcal{S} - \hat{I}_\text{b}$. 
To augment the residuals, vertex-wise colors are sampled from $\mathcal{R}$ via the geometry of a reconstructed 3D face $\mathcal{F}$. These sampled colors are then re-rendered onto another reconstructed 3D face $\mathcal{F}^\mathcal{T}$, randomly chosen from the target identity set $\mathcal{T}$. This process produces the augmented residual $\tilde{\mathcal{R}}$.
Each of the 3,068 makeup images in $\mathcal{S}$ undergoes 100 such augmentations to ensure sufficient diversity.

\subsection{Method}
\label{sec:method}

\begin{figure}[t]
    \centering
    \includegraphics[width=\linewidth]{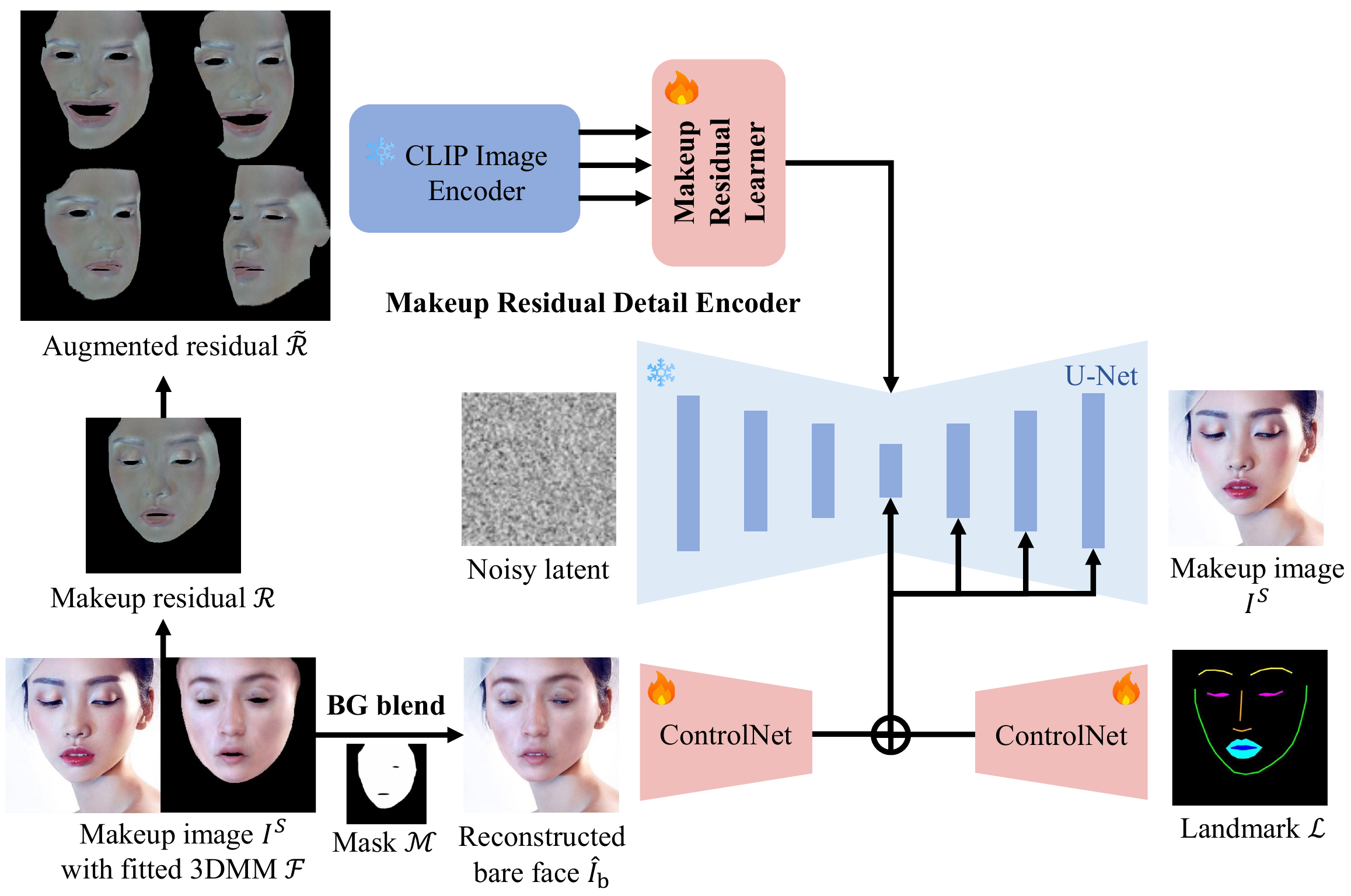}
    \caption{Overview of the FFHQ-Makeup dataset generation method. We extract structure-invariant appearance features from augmented makeup residuals, and guide image synthesis with structural priors to ensure facial consistency.}
\label{fig:method}
\end{figure}

As illustrated in Fig.~\ref{fig:method}, our method adopts Stable-Makeup~\cite{stablemakeup2024} as the backbone for makeup transfer, which is built upon the pre-trained Stable Diffusion model~\cite{stablediffusion2022}. The controllable components are divided into two parts: feature extraction and structural control.

For feature extraction, the Makeup Residual Detail Encoder is designed to extract structure-invariant appearance features from the augmented makeup residual $\tilde{\mathcal{R}}$. This module consists of a frozen CLIP image encoder~\cite{CLIP2021} and a Makeup Residual Learner, which is based on the channel-attention architecture proposed in FreeUV~\cite{freeuv_yang_2025}. This design enables the model to ignore spatial information and selectively capture relevant appearance features. 
For structural control, structural guidance is provided using ControlNet~\cite{controlnet23}, which leverages both the reconstructed bare face $\hat{I}_\text{b}$ and facial landmarks $\mathcal{L}$ to guide the generation process and enhance structural consistency. 
Notably, unlike Stable-Makeup, our approach removes the reliance on paired bare–makeup data during training.
The training setup, including the optimization strategy and hyperparameter configuration, follows that of FreeUV.

\subsection{Dataset Construction}
\label{sec:dataset_construction}

\begin{figure*}[t]
    \centering
    \includegraphics[width=\linewidth]{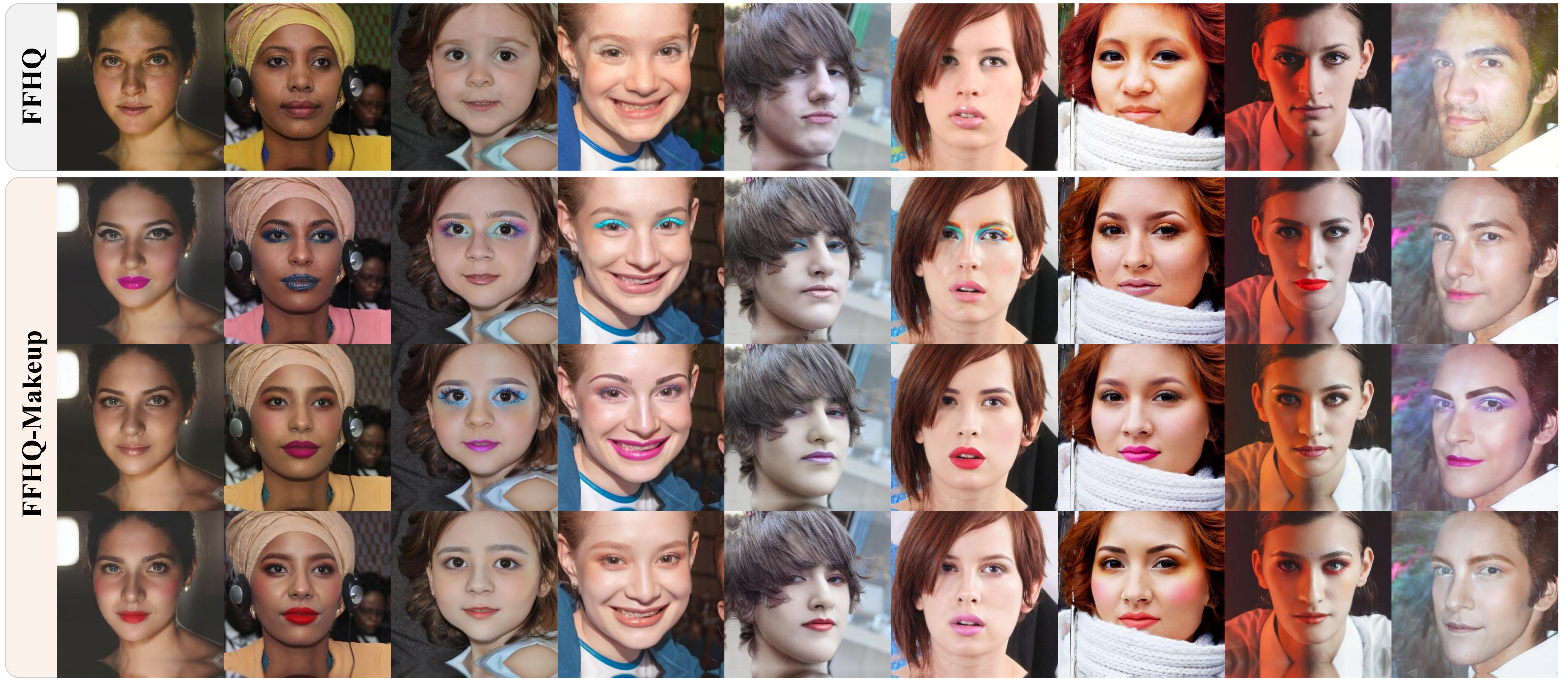}
    \caption{Our FFHQ-Makeup dataset inherits the diversity of FFHQ. As shown, it includes multiple bare-makeup pairs examples across different ethnicities, ages, genders, expressions, and cases with occlusions or shadows.}
\label{fig:result_ffhq_makeup}
\end{figure*}

After training, we apply the trained model for dataset construction. To ensure the quality of the generated dataset, we incorporate careful human inspection and refinement during this phase.

For the makeup source set $\mathcal{S}$ used in appearance extraction, we first remove extreme makeup styles, which are overly rare and may introduce distributional bias. To improve quality, we manually mask out areas in makeup residual $\mathcal{R}$ where facial segmentation fails, particularly in cases with hair overlap or occluded regions. 
After this cleaning process, a curated subset of 2,257 high-quality makeup residuals is retained.

For the target identity set $\mathcal{T}$ from FFHQ used in structural control, we manually filter out samples exhibiting inaccurate 3DMM fitting or failed facial segmentation. For each identity in $\mathcal{T}$, we randomly select 5 makeup styles from the curated makeup source set $\mathcal{S}$ and perform makeup transfer accordingly.

After generation, we conduct a group-wise visual inspection to eliminate samples with artifacts or insufficient visual quality. As a result, the final FFHQ-Makeup dataset comprises 18K unique identities and a total of 90K high-quality images. Examples of the final FFHQ-Makeup dataset are shown in Fig.~\ref{fig:teaser} and \ref{fig:result_ffhq_makeup}.

\section{Evaluation}
We evaluate both the dataset and the data generation method.
For dataset-level quantitative evaluation, we compare our FFHQ-Makeup dataset against existing publicly available large-scale synthetic makeup datasets, including LADN-Syn~\cite{ladn} and BeautyBank~\cite{BeautyBank_Lu_2025}. To ensure a fair comparison, we randomly sample 90K images from each dataset to match the scale of our FFHQ-Makeup.
For generation method evaluation, we conduct an ablation study to assess different input variants and compare against a baseline that directly uses makeup transfer for data generation. In this setting, the evaluation is conducted on a randomly selected subset of 5,000 unfiltered (i.e., not manually cleaned) outputs.

\subsection{Dataset Evaluation}
\begin{figure}[t]
    \centering
    \includegraphics[width=\linewidth]{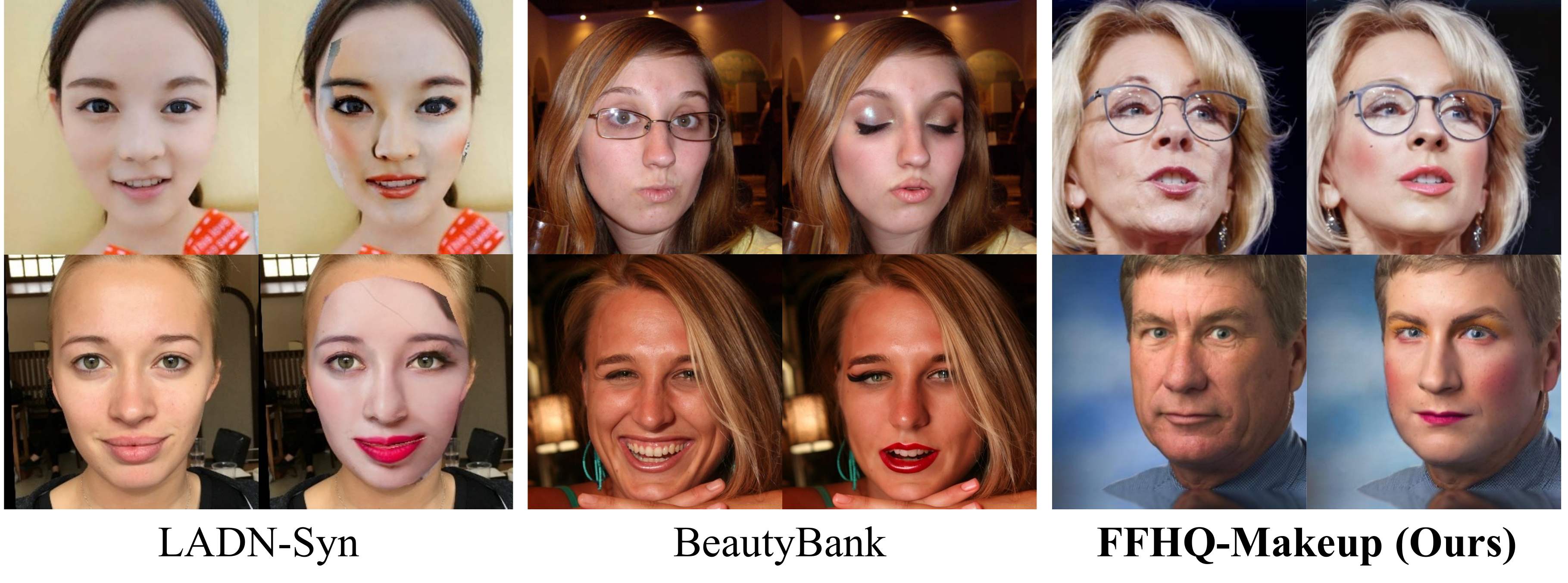}
    \caption{Qualitative comparison of makeup datasets. Our FFHQ-Makeup dataset achieves realistic makeup effects while preserving facial structure and identity.}
\label{fig:result_dataset_evaluation}
\end{figure}

\begin{table}[t]
\centering
\caption{Visual preference results. Scores represent the percentage of times each dataset was selected as best in makeup realism ($\mathcal{P}_\text{makeup}$) and facial consistency ($\mathcal{P}_\text{consistency}$). Our dataset notably outperforms others, especially in maintaining facial consistency.}
\label{tab:dataset_vlm_study}
\begin{tabular}{l|ccc}
\toprule
 & LADN-Syn & BeautyBank & \textbf{Ours} \\
\midrule
$\mathcal{P}_\text{makeup}$ & 0\% & 48\% & 52\% \\
$\mathcal{P}_\text{consistency}$ & 0\% & 8\% & 92\% \\
\bottomrule
\end{tabular}
\end{table}

As shown in Fig.~\ref{fig:dataset_compare} and ~\ref{fig:result_dataset_evaluation}, LADN-Syn constructs its dataset via a warp-and-paste strategy, which directly overlays makeup regions onto target faces. This naive compositing approach often leads to unrealistic appearances and noticeable artifacts. In contrast, BeautyBank employs a text-to-image generation paradigm. However, due to the inherent ambiguity of natural language prompts, the generated results often exhibit unintended changes in identity and facial expression.
Our method, by comparison, produces realistic bare-makeup face pairs while preserving facial structure and identity.

To evaluate dataset quality, we select 50 groups of paired samples and conduct a visual preference study using a vision-language model (GPT-4o~\cite{gpt4o}). The evaluation considers two criteria: makeup realism ($\mathcal{P}_\text{makeup}$) and facial consistency ($\mathcal{P}_\text{consistency}$), with their definitions.
We use the following prompt for the evaluation: "\textit{The following three sets of images are from different makeup datasets. Please select the one you consider the best in terms of Makeup Realism, and Facial Consistency.}"
As shown in Tab.~\ref{tab:dataset_vlm_study}, our approach achieves the highest scores, especially in facial consistency ($\mathcal{P}_\text{consistency}$) across bare–makeup pairs.

\subsection{Ablation Studies}
\begin{figure}[t]
    \centering
    \includegraphics[width=\linewidth]{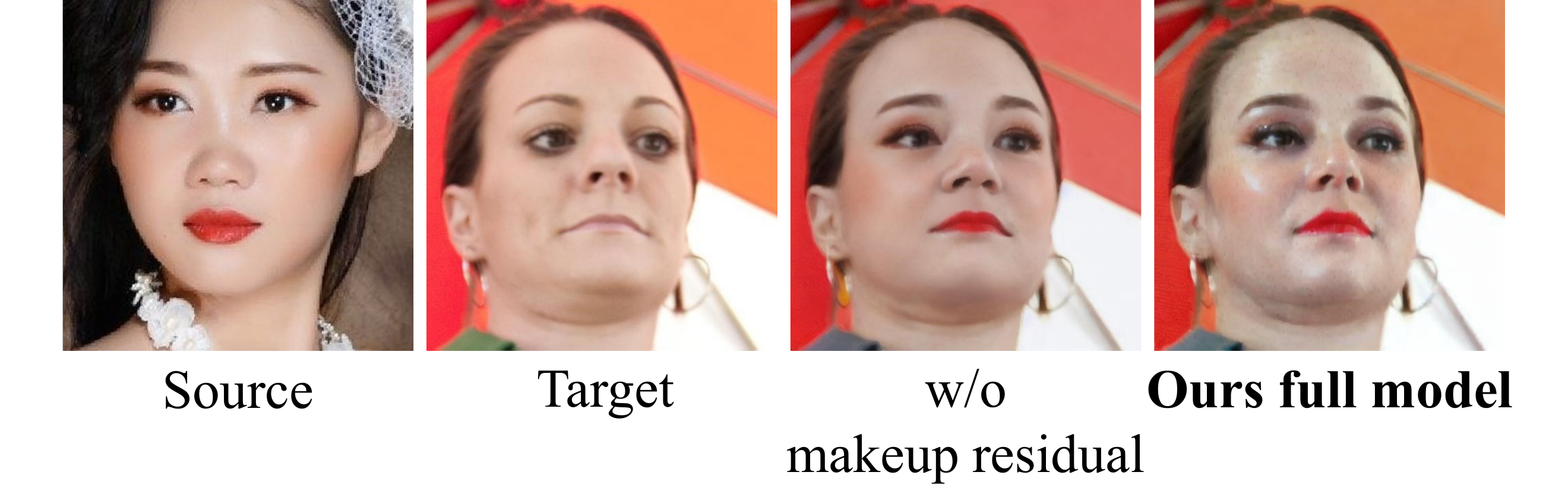}
    \caption{Ablation study on feature extraction without using makeup residual. Feeding the full makeup image causes identity leakage and entanglement between makeup style and source facial features.}
\label{fig:result_ablation_wo_residual}
\end{figure}

\begin{figure}[t]
    \centering
    \includegraphics[width=\linewidth]{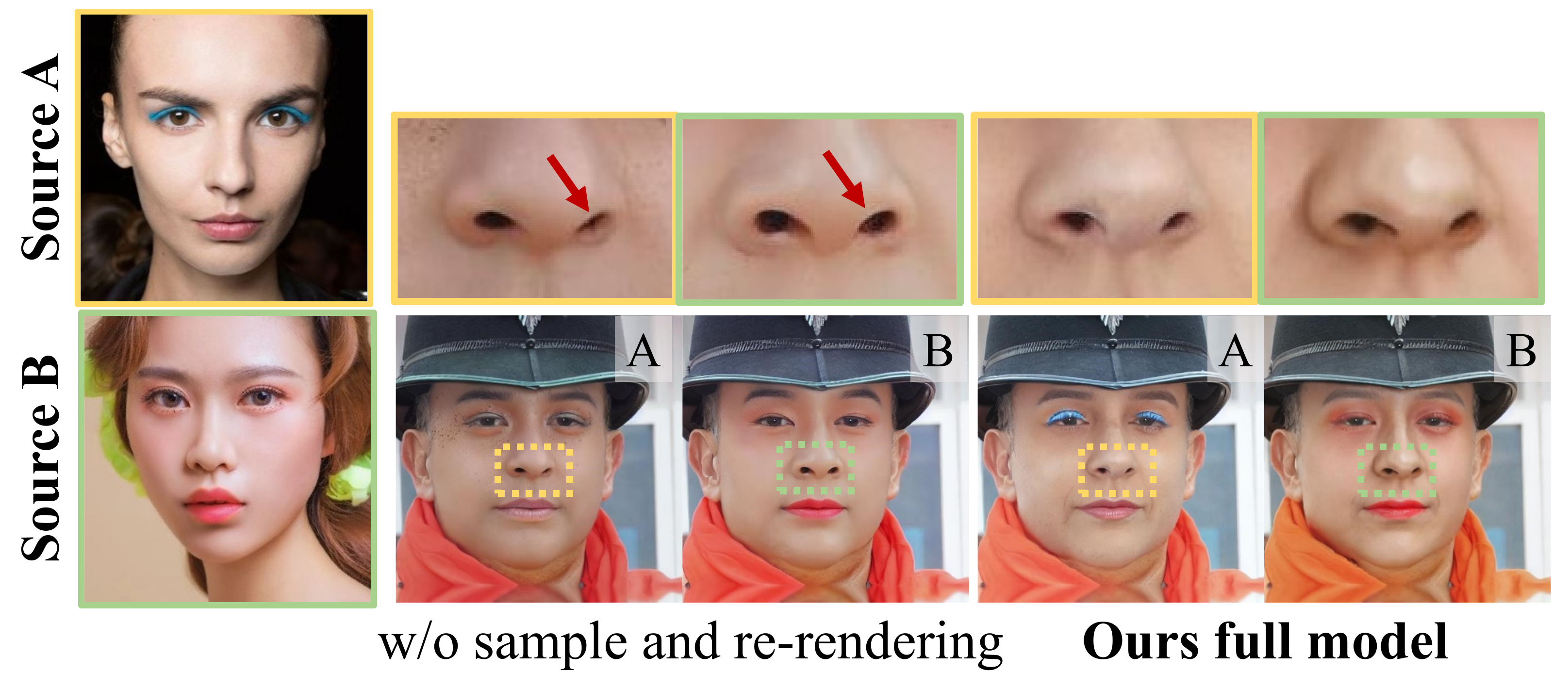}
    \caption{Ablation study on feature extraction without sampling and re-rendering augmentation. Direct use of raw residual leads to structural artifacts leaking from the source face into the generated results.}
\label{fig:result_ablation_wo_aug}
\end{figure}

We compare two ablated training variants to validate the effectiveness of our residual-based representation and augmentation strategy.

The first variant (w/o makeup residual) directly feeds the makeup image $I^\mathcal{S}$ into the feature extraction module, instead of using the residual representation 
$\mathcal{R}$ or $\tilde{\mathcal{R}}$. 
This setting is similar to Stable-Makeup~\cite{stablemakeup2024}, where the source image inherently contains both makeup and identity features. As shown in Fig.~\ref{fig:result_ablation_wo_residual}, this approach leads to entanglement between source identity and makeup, causing identity leakage: the transferred result reflects both makeup style and facial features of the source, thereby compromising identity preservation of the target. In contrast, our residual is derived by subtracting a self-reconstructed face (via 3DMM) from the original, effectively suppressing identity cues and achieving better disentanglement.

The second variant (w/o sample and re-rendering) uses the raw residual $\mathcal{R}$ directly without applying the proposed sampling and re-rendering augmentation. As illustrated in Fig.~\ref{fig:result_ablation_wo_aug}, this leads to residual artifacts where subtle structural traits of the source face (e.g., the shape of nostrils) are still present in the output. Our full model mitigates this by re-rendering the residual on diverse FFHQ geometries, encouraging the network to focus purely on makeup-related features and remain invariant to facial structure.
\textbf{}
\subsection{Makeup Transfer for Dataset Generation}
\begin{figure*}[t]
    \centering
    \includegraphics[width=\linewidth]{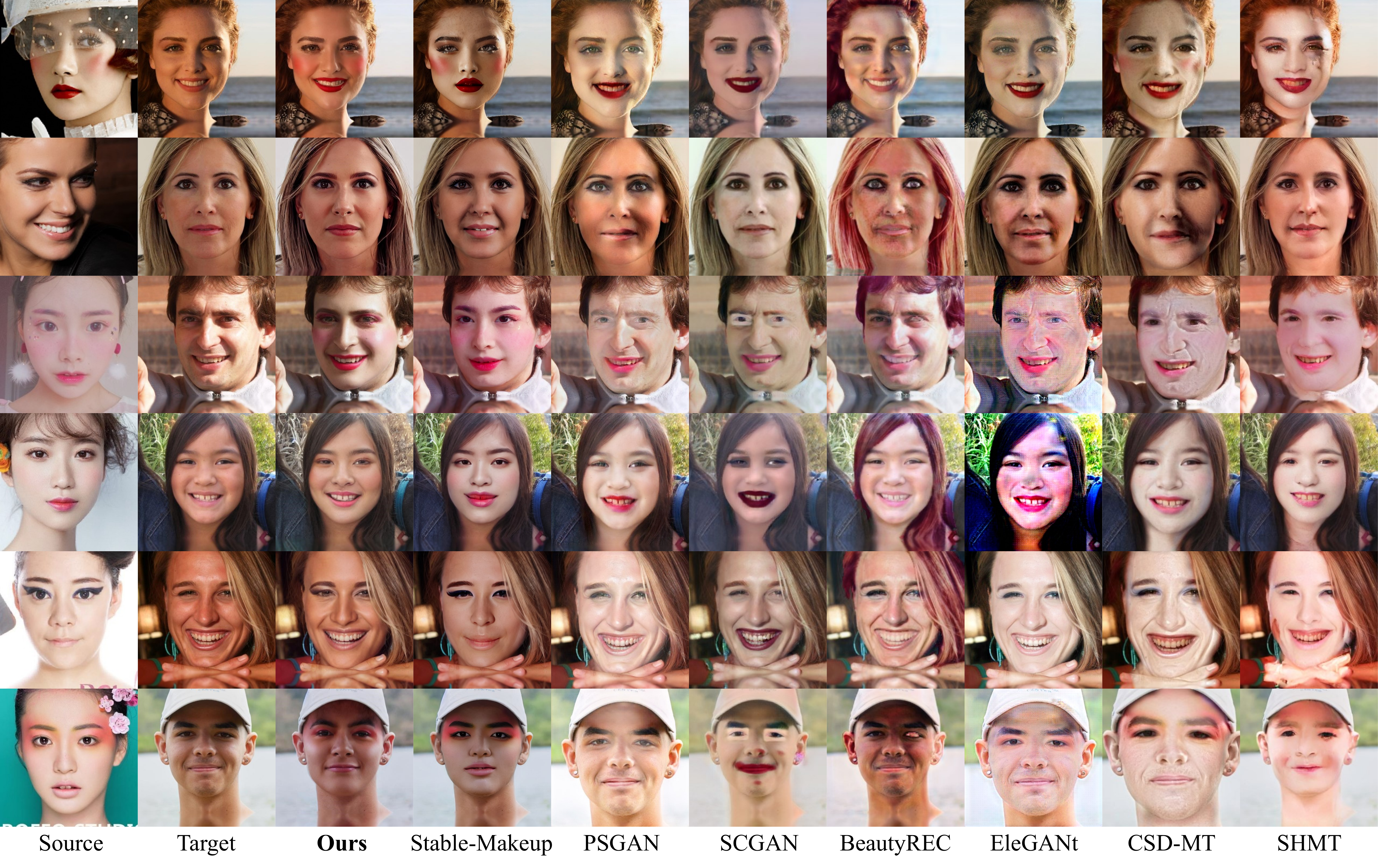}
    \caption{Comparison of makeup transfer methods for dataset generation. Our method best preserves the target identity and expression while producing visually plausible makeup. In contrast, other methods often introduce artifacts or alter key facial attributes such as identity and expression.}
\label{fig:result_makeup_transfer}
\end{figure*}

Our makeup dataset generation method is built upon the Stable-Makeup~\cite{stablemakeup2024} as a baseline, with several improvements introduced to enhance quality and structural control. To validate the effectiveness of our method, we compare it against direct dataset generation using existing makeup transfer methods. Specifically, we evaluate GAN-based methods including PSGAN~\cite{PSGAN}, SCGAN~\cite{SCGAN}, BeautyREC~\cite{BeautyREC}, EleGANt~\cite{elegant}, and CSD-MT~\cite{CSD-MT2024}, as well as diffusion-based methods such as SHMT~\cite{SHMT2024} and Stable-Makeup~\cite{stablemakeup2024}.

As shown in Fig.~\ref{fig:result_makeup_transfer}, the GAN-based methods often produce unstable results with significant visual artifacts. While the baseline method Stable-Makeup effectively captures source makeup appearance, it often alters the target's identity and expression, as it relies on paired training data without mechanisms to properly disentangle appearance from structure.
In contrast, our method achieves the best preservation of target identity and expression, while generating a makeup style that closely resembles the source in a perceptually plausible manner.

\subsection{Quantitative Evaluation}
\begin{table}[t]
\centering
\caption{Quantitative comparison of bare–makeup pairs across different dataset and methods.
Both our dataset and method achieve the highest scores in identity similarity and semantic consistency, highlighting their superior ability to preserve facial consistency.
\textbf{Bolded} values represent the best performance.}
\label{tab:quantitative_comparison}
\begin{tabular}{l|ccc}
\toprule
 & Id~$\uparrow$ & DINO-I~$\uparrow$ & SSIM~$\uparrow$ \\
\midrule
LADN-Syn & 0.4973 & 0.9163 & \textbf{0.9173} \\
BeautyBank & 0.5034 & 0.9008 & 0.8060 \\
\textbf{FFHQ-Makeup} & \textbf{0.5888} & \textbf{0.9448} & 0.8371 \\
\midrule
Stable-Makeup & 0.5191 & 0.9010 & \textbf{0.8431} \\
w/o makeup residual & 0.5346 & 0.9053 & 0.8104 \\
w/o augmentation & 0.5586 & 0.8983 & 0.8126 \\
\textbf{Ours full model} & \textbf{0.5743} & \textbf{0.9251} & 0.8206 \\
\bottomrule
\end{tabular}
\end{table}

We focus the evaluation on the effectiveness of bare–makeup image pairs in maintaining Facial Consistency ($\mathcal{P}_\text{consistency}$) using the following metrics. 
ArcFace~\cite{arcface} is adopted to measure identity similarity between bare and makeup images, reflecting whether both faces belong to the same person.
DINO-I~\cite{DINO2021} is employed to assess high-level semantic consistency, such as pose and expression preservation.
SSIM is included as a low-level structural similarity metric to provide additional insights into pixel-level geometry consistency.

The proposed dataset and its underlying generation method demonstrate superior performance in both identity preservation and semantic structural consistency, validating the effectiveness of our full model in producing high-quality bare–makeup pairs with strong facial consistency ($\mathcal{P}_\text{consistency}$).

\section{Conclusion and Future Work}
We presented FFHQ-Makeup, a new large-scale multiple paired bare–makeup dataset with over 90K image pairs, designed to better balance makeup realism, facial diversity, and facial consistency. Each identity is paired with multiple makeup styles, enabling more diverse and robust usage scenarios. To generate high-quality data without relying on paired training data, we introduce a structure-aware diffusion framework that disentangles identity from makeup using 3DMM-guided residuals and facial re-rendering augmentation. Extensive evaluations show that FFHQ-Makeup outperforms existing datasets in both visual quality and facial consistency. We hope this dataset will facilitate future research on realistic and controllable facial makeup applications.

\subsection{Limitation and Future Work}
While our method achieves strong makeup realism and facial consistency, it occasionally introduces changes beyond the facial region—for example, variations in clothing color, as shown in the second column of Fig.~\ref{fig:result_ffhq_makeup}. Additionally, our results are partially affected by the accuracy of 3DMM fitting and facial segmentation.
Moreover, the diversity of makeup styles is currently limited to those found in the MT~\cite{beautyGAN} and LADN~\cite{ladn} datasets, and facial attributes are confined to the FFHQ~\cite{FFHQ} identity space. In future work, we plan to expand the variety of makeup sources and adopt more diverse face datasets. Currently, the final dataset is manually filtered for quality control; we aim to incorporate automatic evaluation metrics to efficiently scale up the data generation process.

\clearpage
{
    \small
    \bibliographystyle{ieeenat_fullname}
    \bibliography{main}
}

\end{document}